\pdfoutput=1

\documentclass[11pt]{article}

\usepackage[]{EMNLP2022}

\usepackage{times}
\usepackage{latexsym}

\usepackage[T1]{fontenc}

\usepackage[utf8]{inputenc}

\usepackage{microtype}

\usepackage{inconsolata}

\usepackage[flushleft]{threeparttable} 
\usepackage{kotex}
\usepackage{multicol}
\usepackage{booktabs}
\usepackage{makecell}
\usepackage{graphicx}

\newcommand{\embz}{\textsc{Embezzlement}}
\newcommand{\fraud}{\textsc{Fraud}}

\newcommand{\drunk}{\textsc{Drunk driving}}
\newcommand{\rcriminal}{\textsc{Ruling-criminal}}

\newcommand{\lcubebzero}{\textsc{GPT2}-base} 
\newcommand{\lcubeb}{\textsc{LCube}-base} 

\newcommand{\ours}{\textsc{Isla}} 
\newcommand{\Ours}{\textsc{end-to-end Information extractor for Statistical Legal Analysis}} 

\newcommand{\lcubem}{\textsc{LCube}-medium} 

\title{Data-efficient End-to-end Information Extraction \\for Statistical Legal Analysis}

\author{
Wonseok Hwang$^{a}$ \quad 
Saehee Eom$^{a}$ \quad 
Hanuhl Lee$^{a}$ \quad
Hai Jin Park$^{b}$ \quad 
Minjoon Seo$^{a,c}$\\
  $^a$LBox \quad $^b$ Hanyang Univ. \quad $^c$KAIST\\
  \texttt{wonseok.hwang@lbox.kr} \quad 
  \texttt{saeheeeom99@lbox.kr} \quad 
  \texttt{leehanuhl@lbox.kr} \\
  \texttt{haijinpark@hanyang.ac.kr} \quad
  \texttt{minjoon@kaist.ac.kr}\\
}
\begin{document}
\maketitle
\begin{abstract}
Legal practitioners often face a vast amount of documents.
Lawyers, for instance, search for appropriate precedents favorable to their clients, while the number of legal precedents is ever-growing. 
Although legal search engines can assist finding individual target documents and narrowing down the number of candidates, 
retrieved information is often presented as unstructured text and users have to examine each document thoroughly which could lead to information overloading. 
This also makes their statistical analysis challenging.
Here, we present an end-to-end information extraction (IE) system for legal documents. 
By formulating IE as a generation task, our system can be easily applied to various tasks without domain-specific engineering effort.
The experimental results of four IE tasks on Korean precedents shows that our IE system can achieve competent scores (-2.3 on average) compared to the rule-based baseline with as few as 50 training examples per task and higher score (+5.4 on average) with 200 examples. 
Finally, our statistical analysis on two case categories --- drunk driving and fraud --- with 35k precedents reveals the resulting structured information from our IE system faithfully reflects the macroscopic features of Korean legal system.
\end{abstract}

\section{Introduction}
Legal practitioners often need to analyze a vast number of documents while preparing legal cases.
For instance, finding appropriate precedents from ever-growing court decisions on previous cases can be challenging due to its number, while they are critical for decision making on subsequent legal actions.

Although legal search engines based on both lexical and semantic similarity can dramatically decrease the burden, retrieved texts are still unstructured and require further reading. Furthermore, statistical analysis of legal documents is impossible without additional structuralization. 
Such statistical analysis from a vast amount of documents, if possible, may help decreasing implicit bias on judicial decision \citep{levinson2017bias}.

However, structuralizng legal documents is very challenging due to their diversity as they reflect virtually all social phenomena.
This makes building a comprehensive ontology and IE system very demanding.
Instead of building a single perfect complex ontology and corresponding IE system, one can resort to building a task-specific IE system focusing on small number of relevant target information at each task. However, the number of required IE systems will quickly grow together with their development and maintenance cost due to the need of task-specific engineering. The cost from the task- and domain-specific engineering can be reduced if end-to-end systems based on generative models are employed but such systems are often unstable and require a large amount of training data. 

In this study, we try to answer the following questions. (1) Is it possible to build an end-to-end (generative) neural system for legal information extraction showing high precision? If so, (2) how many training examples will be required? (3) What would be the best model architectures for this? (4) Would prompt-tuning be more efficient than fine-tuning? To answer these questions, we perform experiments using various language models with different model sizes and architectures on four IE tasks on Korean precedents. We show that the resulting end-to-end system (\ours\footnote{\Ours}) can achieve competent or better performance compared to the rule-based baseline with only 50 training examples per task. With 200 training examples, \ours\ achieves up to +27\% $F_1$ compared to the baseline. 
Using \ours, we, for the first time, perform a large scale statistical analysis of ``drunk driving'' (24k samples) and ``fraud'' (11k samples) cases from Korean criminal trials. The results show that the structured information by our IE system faithfully reflects the macroscopic features of the Korean legal system.

Our contribution can be summarized as below.
\begin{itemize}
    \item We show that an end-to-end IE system for statistical legal analysis can achieve competent performance compared to the rule-based baseline with as few as 50 training examples.
    \item We also show the result of large-scale statistical legal analysis by structuralizing the data using our IE system on two criminal categories: ``drunk driving'' and ``fraud''.
\end{itemize}

\section{Related Works}
Many previous methods on legal IE tasks are based on tagging (classification) approach \citep{cardellino2017eacl_nerc_ontology,mistica2020ie_sentence_classification,hendrycks2021cuad,habernal2022argument_mining,chen2020-joint-entity-relation,pham2021nllp_legal_termolator,hong2021nllu_ie_dialogue,yao2022FACLlevenEventDetection}. The brief description of individual works are presented in Appendix \ref{sec: previous_tagging_models}.
Compared to these studies, we map all IE tasks into a text-to-text format \cite{raffle2019t5}.
This end-to-end approach removes the burden of task- and domain-specific engineering and is known to show competent or better accuracy compared to tagging based method with enough amount of training examples \citep{hwang2021wyvern, kim2021donut}.
\citet{pires2022s2s_ie} also develop end-to-end IE system to extract information from four types of Portuguese legal documents. 
Compared to this study, we focus on building data-efficient end-to-end system combining both prompt- and fine-tuning methods. In this context, we investigate the effect of scaling model size, pre-training corpus, and training examples. 
Domain-wise, we focus on Korean precedents exclusively. 
Lastly, we show how end-to-end IE system can be applied for statistical legal analysis by analyzing 35k Korean precedents.

\begingroup
\setlength{\tabcolsep}{3pt} 
\renewcommand{\arraystretch}{1} 
\begin{table*}[]
\scriptsize
  \caption{Comparison of various models. The $F_1$ scores of individual fields are shown; BAC (blood alcohol level), Dist (travel distance), Vehicle (type of the vehicle), Rec (previous criminal record on drunk driving), Loss, Loss-A (losses from aiding and abetting), Fine (amount of fine), Imp (imprisonment type and period), Susp (suspension of execution period), Educ (education period), Comm (community service period). The average scores over all tasks are shown in the 5th  column (AVG). 
  All scores are computed using the test sets that consists of 100 examples per task (total 400 examples). 
  }
  \label{tbl_comp}
  \centering
  \begin{threeparttable}
  \begin{tabular}{lcccc|cccc|c|cc|ccccc}
    \toprule
    \multicolumn{1}{c}{Name} &
    \multicolumn{1}{c}{\makecell{Size}} &
    \multicolumn{1}{c}{\makecell{Legal corpus\\size}} &
    \multicolumn{1}{c}{\makecell{\# of training\\examples}} &
    \multicolumn{1}{c}{AVG} &
    \multicolumn{4}{c}{\makecell{\drunk}} & 
    \multicolumn{1}{c}{\makecell{\textsc{Embz}}} & 
    \multicolumn{2}{c}{\makecell{\fraud}} & 
    \multicolumn{5}{c}{\makecell{\rcriminal}}   

    \\
      &  & (tokens)  & (per task)
      & 
      & BAC & Dist & Vehicle & Rec 
      & Loss & Loss & Loss-A 
      & Fine & Imp & Susp & Educ & Comm 
      \\
    \midrule
    \midrule
    \lcubebzero\ (custom) & 124M & 0  & 50
    & 73.0
    & 96.9  &  83.7 & 87.6 & 85.7 
    & 60.1 & 35.9& 0   
    & 80.7 & 93.4 & 98.9 & 92.6 & 60.0 
    
    \\
    mt5-small & 300M & 0  & 50
    & 69.9 
    & 95.8 & 93.0  & 95.7 & 90.1 
    & 72.2 & 42.9   &  0
    & 79.4 & 89.4 & 85.7 & 60.4 & 34.1 
    
    \\
    mt5-large & 1.2B & 0  & 50
    & 74.5 
    & 98.0 & 96.4  & 94.7 & 93.6 
    & 87.5 & 64.8   &  0
    & 84.7 & 82.1 & 96.7 & 68.1 & 27.0
    
    \\
    \midrule 
    \lcubeb$^\mathsection$         & 124M & 259M  & 50
    & 78.6 
    &  99.0  & 88.9 & 90.1 & 95.3 
    & 75.0 & 56.1 & 0
    & 84.7 & 94.2 & 98.9 & 94.7 & 66.7
    
    \\
    \lcubeb~(p)$^\dagger$  & 124M & 259M  & 50
    &  78.5
    & 99.0  & 88.9 & 92.5 & 96.9 
    & 71.0 & 51.2 & 11.1
    & 86.7 & 94.2 & 97.8 & 89.3 & 63.6
     
    \\
    \lcubem & 354M & 259M  & 50
    & 78.6
    & 98.0  & 90.7 &  93.6 & 94.2
    & 73.9 & 58.4 & 0
    & 84.7 & 91.9 & 98.9 & 75.6 & 82.8
    
    \\
    \lcubem~(p) & 354M & 259M  & 50
    &  79.5
    & 98.5  & 91.3 &  93.0 & 96.4
    & 78.3 & 59.0 & 0
    & 86.7 & 92.6 & 98.9 & 92.9 & 66.7
    
    \\
    mt5-small + d.a.         & 300M & 259M  & 50
    & 77.2
    & 99.5  & 93.6 & 94.7 & 94.7
    & 78.5 & 64.8 & 0
    & 77.2 & 92.6 & 98.9 & 90.1 & 41.2
    
    \\
    mt5-small + d.a.~(p) & 300M & 259M  & 50
    &  76.6
    & 98.5  & 91.9 & 94.2 & 93.6    
    & 74.7 & 52.8 & 0
    & 80.7 & 93.4 & 98.9 & 90.8 & 50.0
    
    \\
    \midrule 
    \ours-50$^\ddagger$ & 1.2B & $\sim$1B$^\ast$  & 50
    &  80.5
    & 99.5  & 94.7 & 95.8 & 90.0
    & 90.0 & 69.0 & 10.0
    & 88.9 & 94.2 & 98.9 & 89.7 & 45.7
    
    \\    
    \ours-200 & 1.2B & $\sim$1B$^\ast$  & 200 
    &  88.2
    & 99.5  & 98.0 & 98.9 & 97.4
    & 93.0 & 77.9 & 60.0
    & 92.3 & 95.7 & 98.9 & 94.5 & 52.6
    
    \\    
    \ours\ & 1.2B & $\sim$1B$^\ast$ & --1,000$^\ast$
    & 93.1
    &  99.5 & 97.4 &99.5 &99.0 
    & 91.7 & 80.3& 69.6
    & 95.5 & 95.7 & 98.9 & 98.2 & 92.3 
    
    \\    
    \midrule 
    Rule-based  & - & - & -
    &  82.8
    & 98.0 & 87.6  & 71.8 & 92.5
    & 71.8 & 50.9 & 45.5
    & 88.5 & 97.2 & 98.9 & 98.2 & 92.3
    \\
    \bottomrule
  \end{tabular}
  \begin{tablenotes}[]
  \item $\mathsection$: Our custom GPT2 pre-trained with 150k precedent corpus  \citep{hwang2022lboxopen}.
  \item $^\dagger$: Prompt-tuning.
  \item $\ddagger$: Domain-adapted, and fine-tuned mt5-large with task-specific prompts for individual legal tasks. Our internal datasets were used.
  \item $\ast$: Only the range is shown due to the confidential issue.
  \end{tablenotes}
  \end{threeparttable}
\end{table*}
\endgroup

\section{Tasks} \label{sec: tasks}
We formulate all IE tasks as text generation where a model needs to generate the values of target fields. 
We prepare four IE tasks over Korean precedents, (1) three tasks from the facts and (2) one task from the rulings. Only the precedents from criminal trials of 1st level courts are used.
In the facts IE tasks, a model needs to extract a subset of legally important information from the factual description of cases.
We consider three case categories, (1) \drunk, (2) \embz, and (3) \fraud.
Tasks were selected to be composed of various difficulty levels.
In \drunk, a model extracts ``blood alcohol level'', ``travel distance'', ``type of the vechicle'', and ``previous criminal record on drunk driving''.
In \embz, a model needs to extract the loss from embezzlement.
Finally, in \fraud, a model needs to extract the loss from fraud, or from aiding and abetting fraud.
Although similar to \embz, the facts from \fraud\ cases show more diverse patterns which is reflected in their number of unique words (Table \ref{tbl_data_stat} in Appendix).
In ruling IE task, a model extracts following five fields; (1) amount of fine; (2) imprisonment, (3) suspension of execution, (4) education, and (5) community service periods. 
Examples and data labeling process are shown in Appendix \ref{sec: datasets}.

\section{Models}
We use language models combined with task-specific prompts. 
For a given source text (facts or ruling) and corresponding task-specific prompt, a model generate the values of individual fields autoregressively. See Section \ref{sec: e2e models} for the details.
For the comparison, we develop a rule-based baseline using regular expression. For each of the categories (\drunk, \embz, \fraud and \rcriminal), we read through diverse cases and manually identified suitable identification rules/patterns. See  Section \ref{sec: baseline}.

\section{Experiments}
All experiments are performed on Nvidia A6000 GPU or RTX6000 using Transformers \citep{wolf2020huggin} and  OpenPrompt \citep{ding2021openprompt} libraries. We use \texttt{Precedent corpus} (150k Korean precedents) from LBox Open \citep{hwang2022lboxopen} for the pre-training/domain-adaptation.
See Section \ref{sec: experimental details} in Appendix for the details.

\section{Results}
In this section, we summarize our experimental result on four legal IE tasks with varying difficulties (Section \ref{sec: tasks}). 
We examine 12 models while changing ``model architecture'', ``model size'', ``the size of legal corpus for pre-training'', or ``the size of training dataset'' (Table \ref{tbl_comp}).

\paragraph{End-to-end models can show competent performance with 50 training examples.}
We first examine three language models GPT2-base, mt5-small, and mt5-large.
GPT2-base model shows comparable or lower performance compared to the rule-based baseline on the most of tasks (1st vs final rows) whereas mt5-small and mt5-large shows comparable or better score (2nd, 3rd vs 13th rows). Notably the $F_1$ scores of \fraud\ shows clear improvement upon model scaling and mt5-large (1.2B) exceeds the rule-base baseline.
The low performance on ``damages from aiding and abetting (Loss-A)'', ``education (Educ)'', and ``community service (Comm)''  (1st--3rd rows) are due to lack of enough number of training examples for corresponding fields ( Table \ref{tbl_data_stat}, 1st row in Appendix).

\paragraph{Pre-training with Legal corpus is critical.}
The use of legal corpus greatly improves the accuracy across all tasks for both \lcubebzero\ (+5.6 $F_1$ on average, 1st vs 4th rows, first column) and mt5-small (+7.3 $F_1$ on average, 2nd vs 8th rows).  
    Notably, on Loss field of \fraud\ task, \lcubebzero\ shows +20.2 $F_1$ and mt5-small shows +21.9 $F_1$ (11th column) results highlighting the importance of pre-training with domain-specific corpus for IE tasks.

\paragraph{Prompt tuning does not show clear advantage.}
To investigate whether or not prompt-tuning can improve the model performance compared to fine-tuning approach, we perform control experiments with \lcubeb\ (4th vs 5th rows), \lcubem\ (6th vs 7th rows), and domain-adapted mt5-small (8th vs 9th rows) w/ or w/o fixing the parameters of the language models. 
Interestingly, under our experimental conditions, prompt-tuning does not show clear advantage over fine-tuning on generative IE tasks. Especially on \fraud\ task, $F_1$ changed by -4.9 $F_1$ (\lcubeb), +1.4 (\lcubem), and -12.0 (domain adapted mt5-small).

\paragraph{Scaling model size, pre-training corpus, and training examples is beneficial.}
Since the domain-adapted mt5-small shows top performance on \fraud\ and balanced performance over other tasks (8th row), we choose mt5 as a model architecture and perform scaling experiments.
We first scale model (mt5-small $\rightarrow$ mt5-large) and perform the domain adaptation for 7 epochs with our internal precedent corpus (256M $\rightarrow$ $\sim$1B tokens). The clear improvement over mt5-large (+6.0 $F_1$ on average, 3rd vs 10th rows) and the domain-adapted mt5-small (+3.3 $F_1$ on average, 8th vs 10th rows) are observed, showing the importance of scaling pre-training corpus and model size.
Scaling training examples (50 $\rightarrow$ 200 examples) also leads to +7.7 $F_1$ on average (10th vs 11th rows). Notably, $F_1$ of Loss-A in \fraud\ is increased by +50.0. 
Finally, we further collect the training examples up to $\sim$1,000 and achieve 93.1 $F_1$ on average (12th row), an absolute +10.3 $F_1$ improvement over the baseline (final row).

\section{Analysis} \label{sec: analysis}
In this section, we report the result of statistical legal analysis on two case categories, \drunk\ and \fraud. Using \ours, we first extract information from facts and rulings. For high precision, we control recall rate based on the model confidences that are computed by averaging the probabilities of the generated tokens (Appendix \ref{sec: recall rates}).
We analyze 24,230 confident cases out of 33,554 in \drunk\ task, and 
10,898 confident cases from 15,106 in \fraud\ task. 

\begin{figure}[t]
\centering
\includegraphics[width=0.465\textwidth]{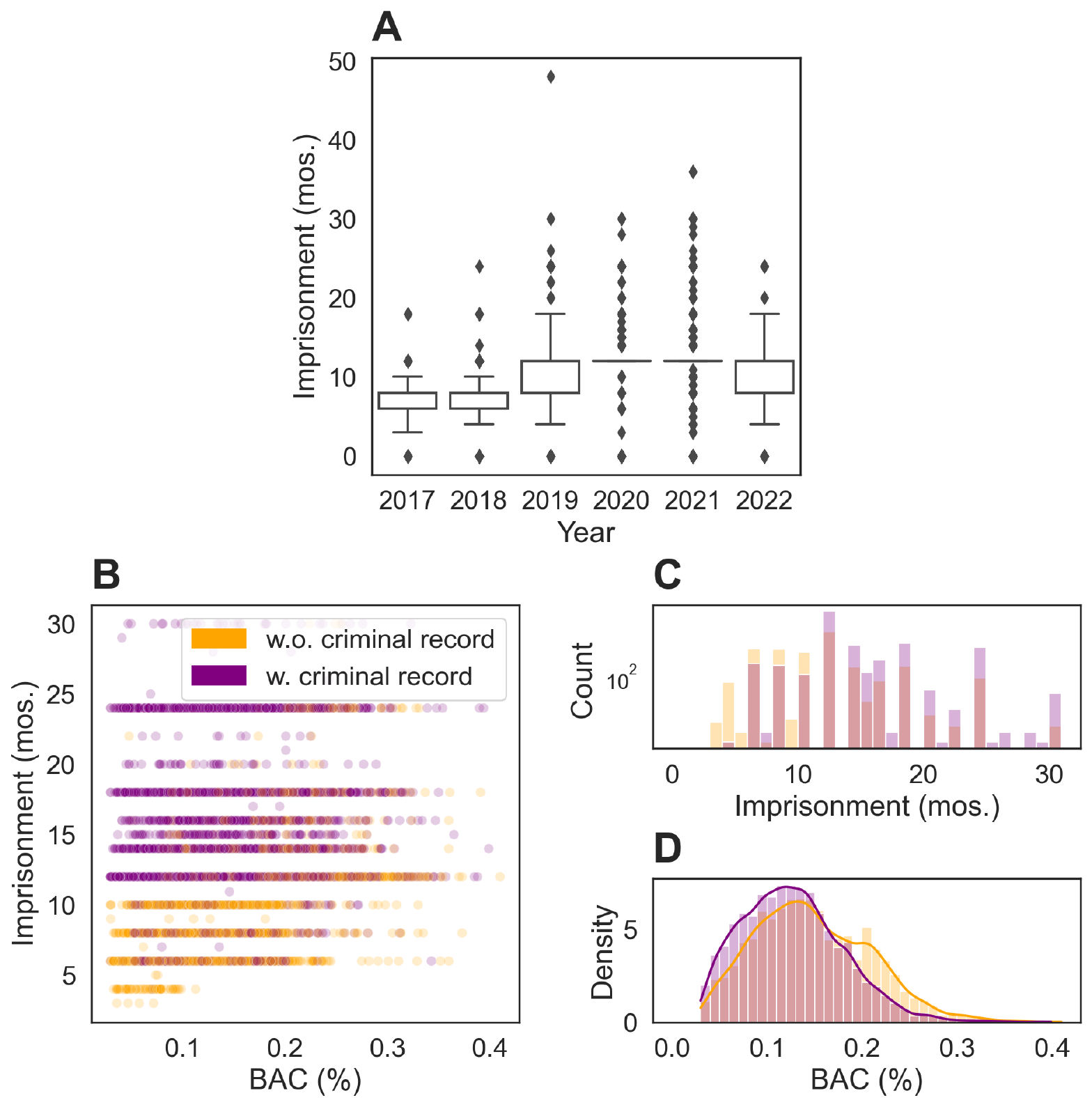}
\caption{
    The result of statistical analysis of \drunk~cases (A: 24k precedents, B--D: 5k precedents). (A) visualizes precedents from 2017 to 2022, and (B--D) visualizes precedents sentenced as suspension of execution or prison after 2019.
    }
\label{fig_la_drunk_driving}

\end{figure}

With the structured data in hand, we first analyze \drunk\ cases (Fig. \ref{fig_la_drunk_driving}).
Two things are noticeable; (1) the average imprisonment period increases since 2019 (A); (2) the people with previous drunk driving record are sentenced longer imprisonment (B, C) regardless of their BACs (D). First result may be related to the fact that the Korean government has strengthened punishment on drunk driving since Jun 25, 2019 
\footnote{Article 148-2 (1) of the Road Traffic Act (amended on December 24, 2018 and went into effect on June 25, 2019; \url{https://elaw.klri.re.kr/kor_service/lawView.do?lang=ENG&hseq=50713)}}.
The second result may reflect the Article 148-2 (1) of the Road Traffic Act which subjects repeat offenders to aggravated punishment\footnote{However, on Nov 25, 2021, Constitutional Court of Korea ruled that the corresponding part of Article 148-2 (1) of the Road Traffic Act (amended on Dec 24, 2018) is unconstitutional (Constitutional Court of Korea 2019Hun-Ba446, 2020Hun-Ka17, 2021Hun-Ba77 (consolidated) ruled on Nov 25, 2021; \url{https://law.go.kr/LSW/detcInfoP.do?mode=1&detcSeq=170177}} (Table \ref{tbl_drunk_driving_avg} in Appendix).

\begin{figure}[t]
\centering
\includegraphics[width=0.465\textwidth]{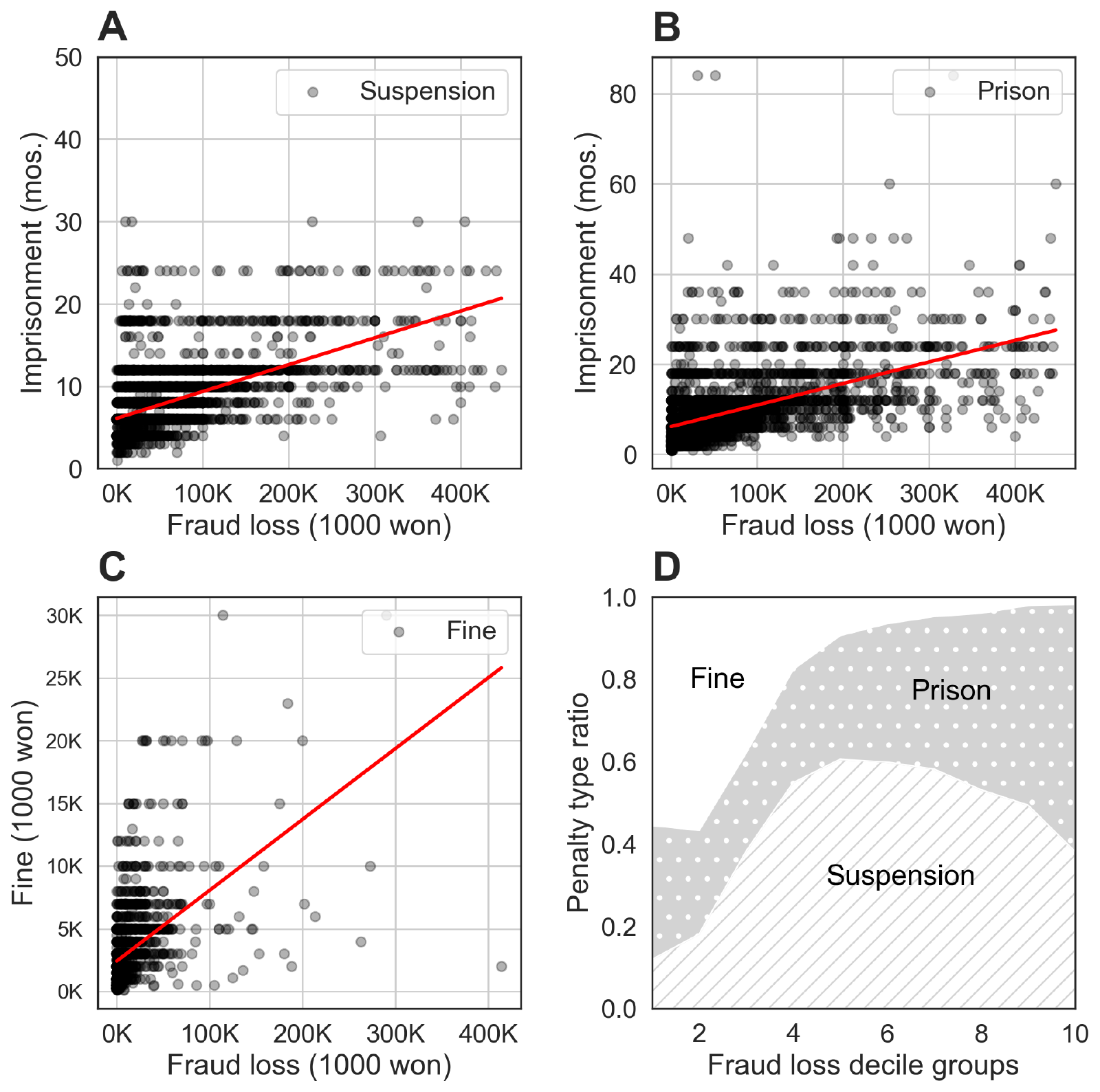}
\caption{
    The result of statistical analysis of \fraud~cases (11k precedents). (A--D) visualizes each record sentenced suspension of execution, prison, and fine as black transparent dots. The regression lines are shown in red.
    }
\label{fig_la_fraud}

\end{figure}

Next we analyze \fraud\ cases. Fig. \ref{fig_la_fraud} shows the increases of imprisonment period and the amount of fine proportional to the loss\footnote{The sentencing guideline for fraud  recommends imprisonment proportional to the magnitude of the loss (\url{https://sc.scourt.go.kr/sc/krsc/criterion/criterion_10/fraud_01.jsp}).
}. Notably, the ratio of cases sentenced as fine decrease with the damages whereas the ratio of cases sentenced as suspension of execution increase initially but decreases with the damages, and the ratio of cases sentenced as prison becomes dominant \footnote{According to Article 62 (1) of the Criminal Act, suspension of sentence can be ruled only when the ruled imprisonment period/fine is not exceeding three years/5 million wons (\url{https://elaw.klri.re.kr/kor_service/lawView.do?hseq=55948&lang=ENG})}.

\section{Conclusion}
We develop a data-efficient end-to-end IE system for legal statistical analysis.
We show that the system can achieve competent performance with small number of examples and exceed the rule-based baseline by large margin upon scaling model size, pre-training corpus, and training examples.
The statistical analysis on 35k precedents reveals the resulting structured data faithfully reflect the macroscopic features of the Korean legal system.

\section*{Ethical considerations}
We present the result of statistical analysis of two legal cases, (1) \drunk\ and (2) \fraud. 
For high precision, we treat only the subset of the precedents in our database (24,230 out of 33,554 in \drunk, 10,898 out of 15,106 in \fraud). Our database also consists of a smaller portion of data compared to the total Korean precedents, as accessing the entire precedents is practically not feasible in Korea due to their purchase cost\citep{hwang2022lboxopen}.
This indicates that our result could emphasize certain properties of the data biasing the result.
Another source of noise is that the model may produce erroneous results on $\sim$3\% samples (Section \ref{sec: analysis}).
Thus, any legal decision based on our analysis should be taken carefully with explicit awareness of above two sources of noises.

\bibliography{legal_ai}
\bibliographystyle{acl_natbib}

\newpage
\onecolumn
\appendix

\section{Appendix}
\subsection{Previous studies on legal IE tasks: tagging-based approaches} \label{sec: previous_tagging_models}
Here we provide brief description of previous tagging methods on legal IE tasks.
\citet{cardellino2017eacl_nerc_ontology} develops BIO-taggers for NER on legal documents. The model is trained with Wikipedia dataset and the result later maps into the legal ontology LKIF based on rules.
\citet{mistica2020ie_sentence_classification} develops the classifier that tags the sentences from Australia precedents into three categories, fact, reasoning, and conclusion. 
\citet{hendrycks2021cuad} presents CUAD, a contract review dataset. In this task, a model needs to extract the span of text and classify in 41 label categories.
\citet{habernal2022argument_mining} develop new argument minding dataset using European Court of Human Rights  with new ontology rooted on legal argument research together with BIO-taggers.
\citet{chen2020-joint-entity-relation} develops a triplet (entities and relation) extraction model on Chinese drug-related criminal judgment documents. Based on BERT encoding, they generate sequence of triplet vectors that are used to tag the position of entities and classify the relation between them.
\citet{pham2021nllp_legal_termolator} develops legal terminology extractor by modifying Termolator that relies on statistical properties of target and background corpora.  
\citet{hong2021nllu_ie_dialogue} develops the IE system that extract 11 features from the dialogue of California parole hearings via classification model.
\citet{yao2022FACLlevenEventDetection} develops large-scale Chinese legal event detection dataset together with various baseline tagging models.

\subsection{Datasets} \label{sec: datasets}

\subsubsection{Data preparation}
We first build the ontologies for the individual tasks (target fields selection) and set-up labeling page using Label Studio \citep{label_studio}.
We place the source text (the facts or the rulings) on the left panel and the workspace on the right.
As all tasks are formulated as text-generation, the workspace consists of simply a list of  ``the name of target field'' and ``text entry box''. The annotators write (often copy and paste) the values of the target field appeared in the source text.
We label 150 randomly selected precedents per each task and split them into 50 training and 100 test examples. 
After that, new examples are added to the training set. 20\% of training examples are used as a validation set.
All datasets were labeled under the guidance of a lawyer.

\subsubsection{Examples} \label{sec: examples}
\subsubsection{\drunk}
\begin{itemize}
    \item Facts: 【범죄전력】 피고인은 2015. 11. 23. 광주지방법원 순천지원에서 도로교통법위반(음주운전)죄로 벌금 400만원의 약식명령을 발령받았다. 【범죄사실】 피고인은 2021. 2. 25. 01:30경 여수시 B모텔 앞 도로에서부터 C에 있는 D 앞 도로에 이르기까지 약 20m 구간에서 혈중알코올농도 0.208\%의 술에 취한 상태로 (차량번호 1 생략) 쏘나타 승용차를 운전하였다. 이로써 피고인은 음주운전 금지 규정을 2회 이상 위반하여 술에 취한 상태로 자동차를 운전하였다."
    \item Label:
    \begin{itemize}
        \item BAC: 0.208\%
        \item Distance: 20m
        \item Vehicle: 쏘나타 승용차
        \item Criminal record: 1
    \end{itemize}
\end{itemize}

\subsubsection{\embz}
\begin{itemize}
    \item Facts:  피고인은 피해자 B이 서울 성북구 C, D호에서 운영하는 E 주식회사에서 1993. 4. 1.경부터 2017. 1. 30.경까지 경리부장으로, 2017. 1. 31.경부터 2017. 10. 31.경까지 사내 이사로 각 근무하면서 임대관리 및 회계관리 등의 업무를 담당하였다. 피고인은 2009. 3. 10.경 위 E 사무실에서 15,596,670원에 대한 지출결의서를 작성하여 피해자의 결재를 득한 다음 같은 날 F은행 동소문로지점에서 위 지출결의서에 따른 예금인출을 하면서 예금청구서의 금액란을 위 지출결의서와 다르게 '19,596,670원'이라고 과다하게 기재한 후 피고인이 업무상 보관중인 E 주식회사 명의의 F은행 금융계좌(계좌번호 : G)에서 '19,596,670원'을 인출한 후 위 지출결의서와의 차액 4,000,000원을 피고인의 개인용도로 임의 사용하여 이를 횡령한 것을 비롯하여 그 무렵부터 2017. 10. 10.경까지 별지 범죄일람표 기재와 같이 총 51회에 걸쳐 합계 502,188,070원을 피해자를 위하여 업무상 보관하던 중 이를 횡령하였다.
    \item Label:
    \begin{itemize}
        \item Loss: [502,188,070원]
    \end{itemize}
\end{itemize}

\subsubsection{\fraud}

\begin{itemize}
    \item Facts:  [2019고정1334] 피고인은 대구 동구 B에 있는 C공인중개사 사무소 직원으로 근무하는 사람이다. 피고인은 2018. 7. 3.경 위 공인중개사 사무소에서 대구 동구 D 원룸 건물주 E으로부터 원룸 세입자를 소개하여 달라는 부탁을 받고 피해자 F에게 '월세 나온게 있는데 월래 월 32만 원인데 계약서상 39만 원으로 적고 월세 39만 원을 내면 차액인 7만 원, 1년치 84만 원을 계약 당일 일시불로 지급하겠다'고 거짓말을 하였다. 그러나 사실 피고인은 피해자가 위 원룸 G호에 대한 월세 계약을 하더라도 대부업체 10여 곳으로부터 5,000만 원 상당의 채무가 있고 지급 능력이 되지 않아 이자도 납부하지 못하고 있어 피해자에게 1년분 월세 차액금 84만 원을 지급할 의사나 능력이 없었다. 그럼에도 불구하고 피고인은 위와 같이 피해자를 기망하여 이에 속은 피해자로부터 대구 동구 D건물 G호에 대한 부동산 월세 계약서를 작성하게 하고 건물주인 E 명의의 계좌로 2018. 7. 3. 실제 월세 32만 원을 초과한 7만 원을 더 많이 송금하게 하는 등 그때부터 같은 방법으로 2018. 8. 5. 7만 원, 2018. 9. 4. 7만 원, 2018. 10. 5. 7만 원, 2018. 11. 5. 7만 원, 2018. 12. 5. 7만 원 등 합계 42만원을 위 E 명의 계좌로 송금하게 하였다. [2019 고정 1335] 피고인은 2018. 9. 중순 일자불상경 대구 동구 H에 있는 I 음식점 안에서 피해자 J에게 '명절을 보내는 데 돈이 필요한데 카드를 빌려주면 2018. 10. 말일경까지 카드대금 변제를 하겠다'는 취지로 거짓말 하였다. 그러나 피고인은 피고인의 명의로 된 재산이 없고, 채무 5,000만 원이 있으나 채무변제도 하지 못하고 있고, 중개보조원으로 수입이 거의 없어서 사실상 피해자의 신용카드를 빌려 사용하더라도 그 대금을 지불할 의사나 능력이 없었다. 피고인은 위와 같이 피해자를 기망하여 이에 속은 피해자로부터 즉석에서 피해자 명의의 신한카드를 건네받아 2018. 9. 21.경 대구 동구 K에 있는 L당구장에서 당구게임 대금 15,000원을 결제한 것을 비롯하여 그 시경부터 2018. 10. 3.경까지 사이에 별지 범죄일람표 기재와 같이 도합 62회에 걸쳐 합계 1,926,934원 상당을 결제 하였으나 카드대금 결제일에 지급하지 않아 피해자로 하여금 그 대금을 대신 지급하게 하여 재산상 이익을 취득하였다.
    \item Label:
    \begin{itemize}
        \item Loss: [42만원, 1,926,934원]
    \end{itemize}
\end{itemize}

\subsubsection{\rcriminal}

\begin{itemize}
    \item Ruling: 피고인을 징역 1년 및 벌금 1,000,000원에 처한다. 피고인이 위 벌금을 납입하지 아니하는 경우 금 50,000원을 1일로 환산한 기간 피고인을 노역장에 유치한다. 다만, 이 판결 확정일로부터 2년간 위 징역형의 집행을 유예한다. 압수된 증 제1호 내지 제3호를 각 몰수한다.
    \item Label:
    \begin{itemize}
        \item Fine: [벌금1,000,000원]
        \item Imprisonment: [징역1년]
        \item Suspension of execution: [2년]
        \item Education: none
        \item Community service: none
    \end{itemize}
\end{itemize}

\subsection{End-to-end IE system} \label{sec: e2e models}
We use decoder-only (GPT2) or encoder-decoder (mt5) language models.
For a given source text (facts or ruling) and following task-specific prompt, the models generate the values of individual fields autoregressively.
For instance, the output of the ruling task looks like ``fine 10,000 won. imprisonment 6 months. suspension of execution 12 months. none. community service 40 hours.'' where the values of multiple fields are generated sequentially separated by "." delimiters. 
The prompts of individual tasks consists of soft tokens (trainable embeddings) initialized from simple declarative sentences  like ``Extract embezzled money.'', ``Write fine, imprisonment, and suspension of execution in sequence''. 
We also perform prompt-tuning experiments where only the soft tokens are trained.
All models are trained with cross-entropy loss from the generated tokens under multi-task setting. The examples are randomly sampled with equal ratio from individual tasks.
All parameters are shared except the soft tokens.
In addition to four IE tasks, a civil ruling IE task is included as an auxiliary task during the training.
In this task, a model extracts ``approved money'' and ``the ratio of litigation cost that plaintiffs' should pay'' from the rulings and ``claimed money'' from the gist of claims. The same number of precedents with other tasks are labeled and used for the training.

\subsection{Rule-based baseline}  \label{sec: baseline}
We develop a rule-based baseline using regular expression for comparison with the end-to-end approaches.
For each of the categories (\drunk, \embz, \fraud and \rcriminal), we read through diverse cases and manually identified suitable identification rules/ patterns.
In \drunk, as ``blood alcohol level'', ``travel distance'' and ``previous criminal record on drunk driving'' appear once each with expressions such as ``\%'', ``km'' or ``more than twice'', these patterns were used to extract each of the information. Also,``type of the vehicle`` was extracted by the pattern ``(drunk)drove <vehicle><object marker>``.
In \embz, the embezzled money is extracted by (1) extracting all monetary values from facts and (2-a) if there is the word ``total`` preceding a monetary value, such amount of money was selected as the total amount of embezzled money. (2-b) Else, we selected the last appearing money. 
In \fraud, the damages are extracted similarly to \embz\ cases except if "aid" appears in the sentence that includes last money, it is considered as loss from fraud aiding and abetting.
In \rcriminal, an amount of fine, imprisonment, suspension of execution, education, and community service periods are extracted by (1) selecting sentences including the indicators such as ``fine'', ``imprisonment'', ``suspension of execution'', ``education'', and ``service'' and (2) extracting numbers with corresponding units (won, years, months, hours, etc) from the same sentence.

\subsection{Experimental details} \label{sec: experimental details}
All models are fine-tuned with batch size 8--16 with learning rate 0.0001 with maximum 100 epochs under multi-task setting (Section \ref{sec: e2e models}). In the prompt-tuning experiments, learning rate is set to 1.0 with maximum epochs 250--300.
GPT-2, \lcubeb, and \lcubem\ models are pre-trained from scratch using Megatron library \citep{shoeybi2019megatron} using \texttt{Precedent corpus} (150k Korean precedents)from LBox Open \citep{hwang2022lboxopen}.
To fine-tune mt5 models, \texttt{google/mt5-small/large} checkpoints are downloaded from Huggingface hub.
For domain adaptation, mt5-small is pre-trained with word level span corruption objective for 22 epochs with the batch size 12.
Also, \ours\ is prepared by pre-training mt5-large starting from the official checkpoint for 7 epochs with batch size 24 using our internal legal corpus.

\subsection{Metric} \label{sec: metric}
In all IE tasks, we calculate $F_1$ as followed. 
True positive if the target field (FLD) exists both in the ground truth (GT) and the prediction (PR) and their values are equal.
False positive if either (1) the values are not equal, or (2) the FLD exsits only in PR.
False negative if the value exists only in GT.
True negative if FLD does not exist in both GT and PR.

\subsection{Recall rates} \label{sec: recall rates}
We set the recall rates to be 84\%, 81\%, and 60\% for \rcriminal, \drunk, and \fraud\ respectively. 
This results in 100\%, 97\%, and 97\% precision on the our internal validation set. 

\subsection{Additional analysis} \label{sec: additional analysis}
\begingroup
\setlength{\tabcolsep}{3pt} 
\renewcommand{\arraystretch}{1} 
\begin{table*}[]
\scriptsize
  \caption{Data statistics.
  }
  \label{tbl_data_stat}
  \centering
  \begin{threeparttable}
  \begin{tabular}{lcccc|cccc|c|cc|ccccc}
    \toprule
    \multicolumn{1}{c}{Name} &
    \multicolumn{1}{c}{\makecell{Size}} &
    \multicolumn{1}{c}{\makecell{Legal corpus\\size}} &
    \multicolumn{1}{c}{\makecell{\# of training\\examples}} &
    \multicolumn{1}{c}{AVG} &
    \multicolumn{4}{c}{\makecell{\drunk}} & 
    \multicolumn{1}{c}{\makecell{\textsc{Embz}}} & 
    \multicolumn{2}{c}{\makecell{\fraud}} & 
    \multicolumn{5}{c}{\makecell{\rcriminal}}   
    
    \\
      
    \toprule
    \# of individual fields & - & - & 50
    &  -
    & 50 & 50 & 50 & 49 & 49 & 48 & 2
    & 19 & 39 & 22 & 10 & 5
    
    \\
    \# of individual fields & - & - & 200
    & -
    & 200 & 194 & 199 & 186 & 196 & 179 & 18
    & 71 & 153 & 101 & 63 & 31
    
    \\
    \midrule 
    \# of unique words & - & - & 50
    & -
    & \multicolumn{4}{c|}{612} 
    & 2,557 & \multicolumn{2}{c|}{4,618} 
    & \multicolumn{5}{c}{286}
    
    \\
    \# of unique words & - & - & 200
    & -
    & \multicolumn{4}{c|}{1,821} 
    & 6,928 & \multicolumn{2}{c|}{11,712} 
    & \multicolumn{5}{c}{664}
    
    \\

    \bottomrule
  \end{tabular}
  \end{threeparttable}
\end{table*}
\endgroup

\begingroup
\setlength{\tabcolsep}{6pt} 
\renewcommand{\arraystretch}{1} 

\begin{table*}[h]
\small
  \caption{The average imprisonment period w/ and w/o criminal records in \drunk\ cases.
  }
  \label{tbl_drunk_driving_avg}
  \centering
  \begin{threeparttable}
  \begin{tabular}{lcc}
    \toprule
    \\
     Year &  w/o criminal record & w/ criminal record
     \\
    \midrule
    \midrule
     2017--2018 & 5.3 months & 7.7 months \\
     2019--2022 & 8.9 months & 11.9  months \\
    \bottomrule
  \end{tabular}
  \end{threeparttable}
\end{table*}
\endgroup

\end{document}